\documentclass[review]{elsarticle}

\usepackage{float}

\usepackage{lineno,hyperref}
\modulolinenumbers[5]

\journal{Journal of \LaTeX\ Templates}

\biboptions{authoryear}

\begin{document}

\begin{frontmatter}

\title{Online Recognition of Actions Involving Objects}

\author[mymainaddress]{Zahra Gharaee\corref{mycorrespondingauthor}}
\cortext[mycorrespondingauthor]{Zahra Gharaee}
\ead{zahra.gharaee@lucs.lu.se}
\author[mymainaddress,Peters_secondaryaddress]{Peter G\"ardenfors}
\ead{peter.gardenfors@lucs.lu.se}
\author[mymainaddress,mysecondaryaddress]{Magnus Johnsson}
\ead{magnus@magnusjohnsson.se}

\address[mymainaddress]{Lund University Cognitive Science, Helgonav\"agen 3, 221 00 Lund, Sweden}
\address[Peters_secondaryaddress]{Social Robotics Studio, Centre of Quantum Computation and Intelligent Systems, University of Technology, Sydney, Australia}
\address[mysecondaryaddress]{Department of Intelligent Cybernetic Systems, NRNU MEPhI, Moscow, Russia}

\begin{abstract}
We present an online system for real time recognition of actions involving objects working in online mode. The system merges two streams of information processing running in parallel. One is carried out by a hierarchical self-organizing map (SOM) system that recognizes the performed actions by analysing the spatial trajectories of the agent's movements. It consists of two layers of SOMs and a custom made supervised neural network. The activation sequences in the first layer SOM represent the sequences of significant postures of the agent during the performance of actions. These activation sequences are subsequently recoded and clustered in the second layer SOM, and then labeled by the activity in the third layer custom made supervised neural network. The second information processing stream is carried out by a second system that determines which object among several in the agent's vicinity the action is applied to. This is achieved by applying a proximity measure. The presented method combines the two information processing streams to determine what action the agent performed and on what object. The action recognition system has been tested with excellent performance.
\end{abstract}


\end{frontmatter}


\section{Introduction}
\noindent Action recognition plays an important role in interactions between any agents whether they are humans, animals or robots. \cite{Johansson} showed by using a patch light technique that humans can identify an action after only about two hundred milliseconds. Such an efficient mechanism for interpreting and categorizing a perceived action is an important factor behind smooth interaction and cooperation between humans. His studies opened up the field of biological motion within psychology.

Later studies of human categorizations of actions have shown a number of features that are relevant also for robotic models. Firstly, categorizations of actions exhibit the same prototype effects as categorizations of objects ( \cite{Hemeren}). Secondly, actions can be categorized in terms of the force patterns involved (\cite{Runesson2},  \cite{Gardenfors2}, \cite{Gardenfors5}). In other words, the dynamics of an action may be more characteristic than its kinematics. Thirdly, human judgments concerning the segmentation of actions show large agreements (\cite{zacks1}).

Given the efficiency of the human action recognition system, it should serve as an inspiration when developing fast and robust methods for action recognition that can be employed in social robotic systems that are interacting with humans online. The general task for such a robotic system is to use online visual data from cameras to track movements of humans and to use this information to categorize actions and then generate an appropriate response, linguistic or non-linguistic. It should be noted, however, that online automatic action recognition is not only useful for human-robot interaction, but also for areas such as video surveillance, human-computer interaction, video retrieval, sign language recognition, medical health care and sport.

In this article we present a biologically inspired system for online action recognition that merges the information analyses from two subsystems running in parallel. To some extent, our architecture is inspired by the two-streams hypothesis about how the brains processes visual information\citep{A.Goodale}. This hypothesis distinguishes between a ventral stream (the ”what” pathway) and a dorsal stream (the ”where” or ”how” pathway). Our two subsystems can be seen as corresponding to these two streams. The first subsystem determines which object the agent acts on by applying a proximity measure (our system, however, takes a shortcut when identifying objects). The second subsystem recognizes what action is performed by using a hierarchical self-organizing map system that analyses the spatial trajectories of the agent's movements.

The fact that the dorsal pathway is called both the ”how” system and the ”where” system reflects that two perspectives can be used when categorizing an action. The first focuses on the manner in which an action is performed (how), for example, whether an object is pushed or pulled. The second perspective focuses on the result of the action, for example, whether an object moves (where) or changes some property. In parallel with this distinction, natural languages contain two types of verbs describing actions \citep{Levin,Warglien}. The first type is manner verbs that describe how an action is performed. In English, some examples are run, swipe, wave, push, and punch. The second type is result verbs that describe the result of actions. In English, some examples are move, heat, clean, enter, and reach. Manner verbs express causes and result verbs express effects of actions. In our previous experiments \citep{Gharaee2, Gharaee3, Gharaee4} actions without objects have been studied and the verbs describing the output have been manner verbs. In the present study that include objects, the output contains both manner and result verbs.

In human-robot applications it is important to collaborate about objects, so it is necessary to develop a system that can categorize actions involving objects as well as pure manner actions. Within robotics, action recognition systems have, until recently, been based on the result perspective, focusing on how result verbs can be modeled, e.g. \citep{Cangelosi,Kalkan,Lallee,Demiris}. From this perspective, it is sufficient to know the pre-state and post-state of the environment before and after performing an action in order to categorize an action. In the method proposed in \cite{Lallee}, the robot learns four actions including objects as "cover", "uncover", "give" and "take" through linguistic interactions with human agents and as a result generates spoken language that represents its perceptions of the performed action. A human-robot communication system that includes both manner and result verbs has been developed in \cite{Mealier}. In this study, the action comprehension and object detection occurs through visual perception (observations) and human-robot spoken interactions (expression of causes and effects of the actions).

 In the literature one finds several systems that can categorize different sets of human actions. In the past, research focused on categorizing actions based on image sequences from ordinary visible light cameras \citep{Niebles}. Unfortunately such cameras have severe limitations such as sensitivity to color and illumination variations, occlusions, and background clutters. As a consequence, Kinect and other depth cameras are often used instead since they provide 3D information about the scene, which offers more discerning information of the human postures involved in the actions that are studied. The depth camera can also operate in total darkness which is a benefit for applications such as continuous patient/animal monitoring systems. The skeletons estimated from depth images are quite accurate, but the algorithm still has limitations. It gives inaccurate results when the human body is partly occluded, and the estimation is not reliable when the person touches the background or when the person is not in an upright position \citep{Xia2}.

In \cite{Li}, a data set of 20 actions, each performed by 10 actors in 2 or 3 different events, has been collected from sequences of depth maps obtained by a depth camera. An action graph is used to model the dynamics of the actions, and a collection of 3D points is used to characterize a set of salient postures corresponding to the nodes in the action graph. The same data set, often called the MSR Action 3D data set, has been studied by many other researchers. Here we briefly present some of the methods that have been used.

In \cite{Xia1}, a method applied to the histogram of 3D joint locations as a compact representation of postures is introduced. It uses Linear Discrimant Analysis to project the histogram of 3D joint locations extracted from the action depth sequences and then clusters them into k posture visual words (the prototypical action poses). Another method for activity recognition from videos gained by a depth sensor is represented in \cite{Oreifej}. It builds a histogram to capture the distribution of the surface normal orientation in the 4D space of time, depth, and spatial coordinates by creating 4D projectors, which quantize the 4D space and represent the possible directions for the 4D normal. The method presented in \cite{Yang-Xiaodong1}, also uses body joints extracted from sequences of depth maps. It applies features based on position differences of joints (eigen joints) that combine action information including static posture, motion, and offset and then uses the Naive Bayes Nearest Neighbour classifier to classify actions. To recognize human actions from depth maps, \cite{Yang-Xiaodong2} use depth maps that are projected onto three orthogonal planes and global activities through entire video sequences that are accumulated to generate a Depth Motion Map. Then the histograms of Oriented Gradients are extracted from the Depth Motion Map as the representation of an action video.

A pose-based action recognition system is introduced in \cite{Wang-Chunyu} that extends the method in \cite{Yang-Yi} to estimate human poses from action videos. It infers the best poses by best-K pose estimation for each frame by incorporating segmentation and temporal constraints for all frames in the video. A visual representation for 3D action recognition from sequences of depth maps, called Space-Time Occupancy Patterns, is used in \cite{Vieira}. In the proposed feature descriptor method, a 4D grid for each depth map sequence is produced by dividing space and time axes into multiple segments to preserve spatial and temporal information between space-time cells. In \cite{Wang2}, semi-local features called Random Occupancy Pattern features are extracted and a sparse coding approach is used to encode these features. In \cite{Wang1}, an actionlet ensemble model which learns to represent each action and to capture the intra-class variance is introduced. The model proposes features of depth data that are capable of characterizing human motion and human-object interactions. The use of local spatio-temporal interest points (STIPs) and the resulting features from RGB videos is the base of the activity recognition method presented in \cite{Xia2}. In the method, first the STIPs are extracted from depth videos (called DSTIP), and then 3D local cuboid in depth videos by Depth Cuboid Similarity Feature (DCSF) are described. Using DSTIP and DCSF to recognize activities from depth videos have no dependence on the skeletal joints information. A non-parametric Moving Pose framework for low-latency human activity recognition is proposed in \cite{Zanfir}, which is a descriptor that considers the pose information together with the speed and acceleration of the skeleton joints. The descriptor works with a modified k-nearest neighbours classifier, which employs both the temporal location of a particular frame within the action sequence as well as the discrimination power of its moving pose descriptor compared to other frames in the training set.

Common to all the systems presented here is that they use a pre-recorded data set of actions, typically the MSR data \citep{Li}. The movies for actions are edited so that they only cover one of the actions that will be categorized and not the intermediate intervals. The systems are then trained to classify the actions. The experiments are mostly performed on one specific way of data assortment. This means that the systems are in general not tested on movies outside the data set and it is unknown to what extent they can generalize to new movies. Moreover, these systems are tested in offline experiments while in human-robot interaction scenarios, the system are supposed to identify actions online in real time. 

Among related studies that propose an online implementation of action recognition, \cite{Ellis} proposes an online action classification method based on the canonical body poses, and a feature descriptor based method for classifying actions from depth sequences introduced in \cite{Vieira}. In both studies, the actions used in online experiments do not involve objects, only body movements which form the agent's spatial trajectories.  

Variants of our system presented below have also been trained and tested with the prerecorded MSR data set in the research studies presented in \citep{Gharaee3, Gharaee4}. In addition to that, our system is also used in online experiments with new actions in \citep{Gharaee2}. 

So far the research studies encountered the problem of action recognition from either of the two perspectives as mentioned earlier. From the perspective that focus on the manner in which an action is performed the spatial trajectory of the actor is the main function used to recognize the actions. While from the other perspective in which the focus is on the results of the action the resulting conditions of the actor and the environment (including the object) is the matter of importance. We propose a method which combines both perspectives in order to have a complete comprehension of actions that helps us to understand and recognize a larger number of actions with a good performance. In our method we analyze the performance of the actions in their context by extracting the information of how an action is performed from the actor's spatial trajectory and what happens as a result of performing that action through the changes in the position of the object. 

The present version of our action recognition system has been implemented in C++ using the neural modelling framework Ikaros \citep{balkenius}. It employs a Kinect sensor to obtain visual input and uses the software libraries OpenNI and NITE to read the sensor and extract the joint positions of detected agents.

The rest of the paper is organized as follows: The architecture is described in section 2, the action recognition experiments to evaluate the architecture are presented in section 3, and section 4 provides a discussion about the approach.

\section{Architecture}
\label{sec:3}
\begin{figure}[!h]
  \centering
   \includegraphics[width=1.00\textwidth]{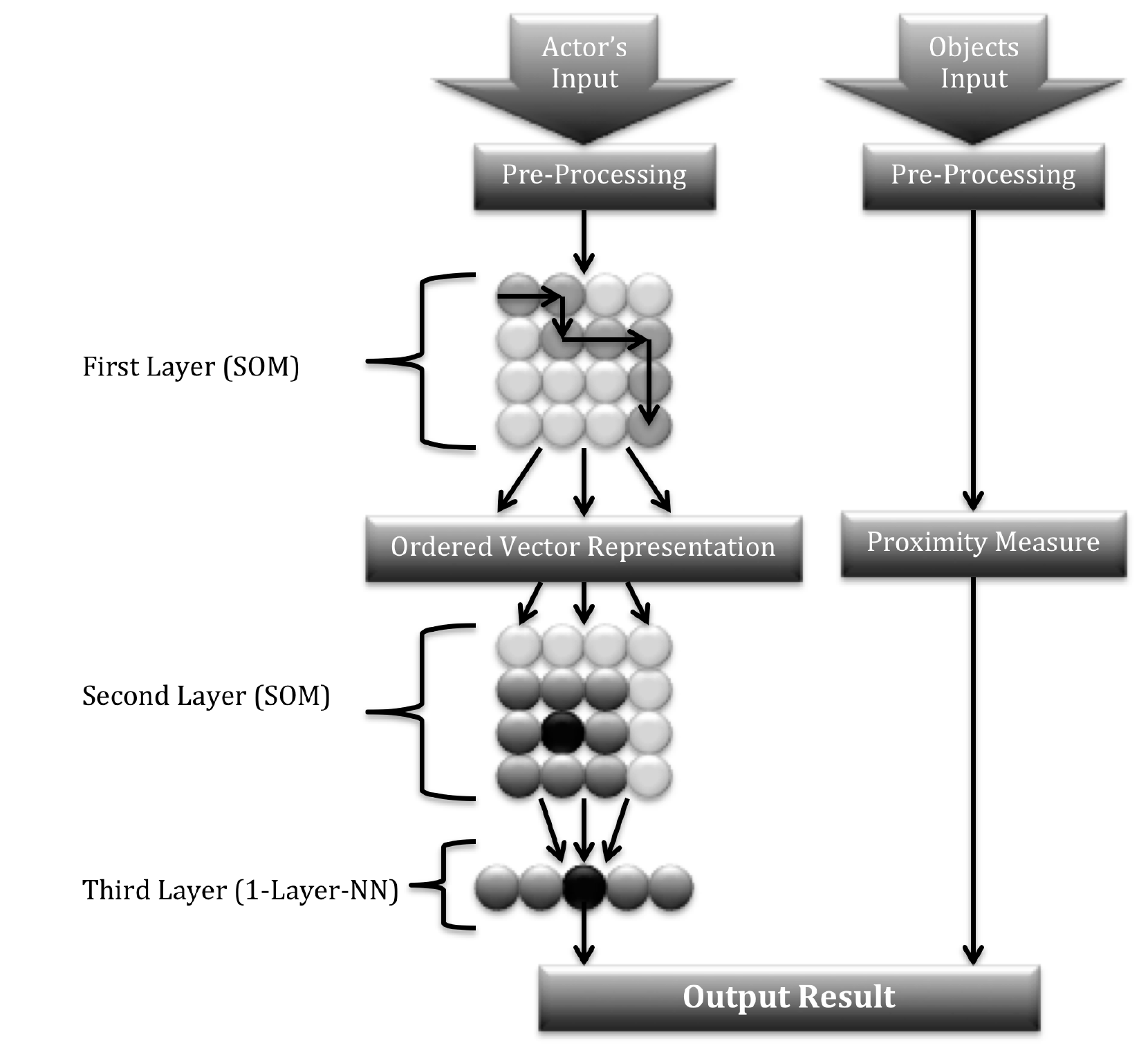}
  \caption{The action recognition architecture that is composed of two subsystems. One of these subsystems (to the left) is a hierarchical neural network system that recognizes the actions carried out by the agent. The first layer consists of a SOM. Layer two consists of a SOM and layer three of a custom made supervised neural network. The darker arrows in the first layer SOM represent the activity trace during an action. The other subsystem (to the right) tracks the present objects and determines the particular object the agent acts upon by applying a proximity measure.}
  \label{fig:HY_HSOM}
 \end{figure}

\noindent The action recognition architecture presented in this article recognizes actions involving objects, Fig.~\ref{fig:HY_HSOM}. This means that in an environment with several objects present, both the specific action carried out by the observed agent, and the particular object it is applied to will be determined. The action recognition architecture employs two subsystems running in parallel, and synthesizes an estimation based on their output. One of the subsystems is a hierarchical SOM based neural network system that recognizes the actions based on the movements of the observed agent. This is done by extracting sequences of postures making up the actions, clustering these sequences, and labelling the clusters. The other subsystem simultaneously detects and tracks the objects that are present. Since object recognition is not the focus of our research, the present implementation bases the object tracking on markers attached to the objects. The implementation is modularized, so this marker tracking could in principle be replaced by a more sophisticated object tracking mechanism that determines the identities of the present objects by object recognition. Action segmentation has not been addressed in our research, so the current implementation of our online action recognition architecture receives signals indicating the start and the end of each action by a manually operated switch while the actions are performed. The object involved in the action is determined by applying a proximity measure between the observed agent's hand and the present objects.

\subsection{Action Recognition}
\label{sec:SOM}
\noindent The action recognition subsystem recognizes the specific action carried out by the agent. It is composed of three neural network layers. The first layer consists of a SOM that develops a compressed and ordered representation of the preprocessed input (i.e. parts of scaled postures in an egocentric framework) obtained by a Kinect sensor. The second layer consists of a second SOM and receives ordered vectors that are spatialized representations of the activity patterns elicited in the first-layer SOM during actions. The ordered vector representation of the sequence of unique activations in the first layer provides a mechanism that makes the action recognition time invariant. This is possible because similar movements carried out at different performance speeds will elicit similar sequences of unique activations in the first layer SOM. Thus the second layer that receives these ordered vector representations will learn to cluster complete actions. The third layer consists of a custom made supervised neural network that labels the activity in the second layer SOM with the corresponding actions. The third layer could provide some independence from the viewing angle of and the distance to the camera, but this is done more efficiently as a part of the pre-processing, i.e. by scaling and transforming the sets of joint positions into an egocentric framework before they are received by the first layer SOM, as will be explained next.\\

\noindent \textbf{Pre-processing.} The action recognition subsystem uses a stream of sets of joint positions obtained by preprocessing the original stream of depth images from a Kinect sensor. This preprocessing is accomplished by code relying on the software libraries OpenNI and NITE for reading the Kinect sensor and extracting the sets of joint positions. Each such set of joint positions contains 15 joint positions expressed in 3D Cartesian coordinates.

\begin{figure}[!h]
  \centering
   \includegraphics[width=0.80\textwidth]{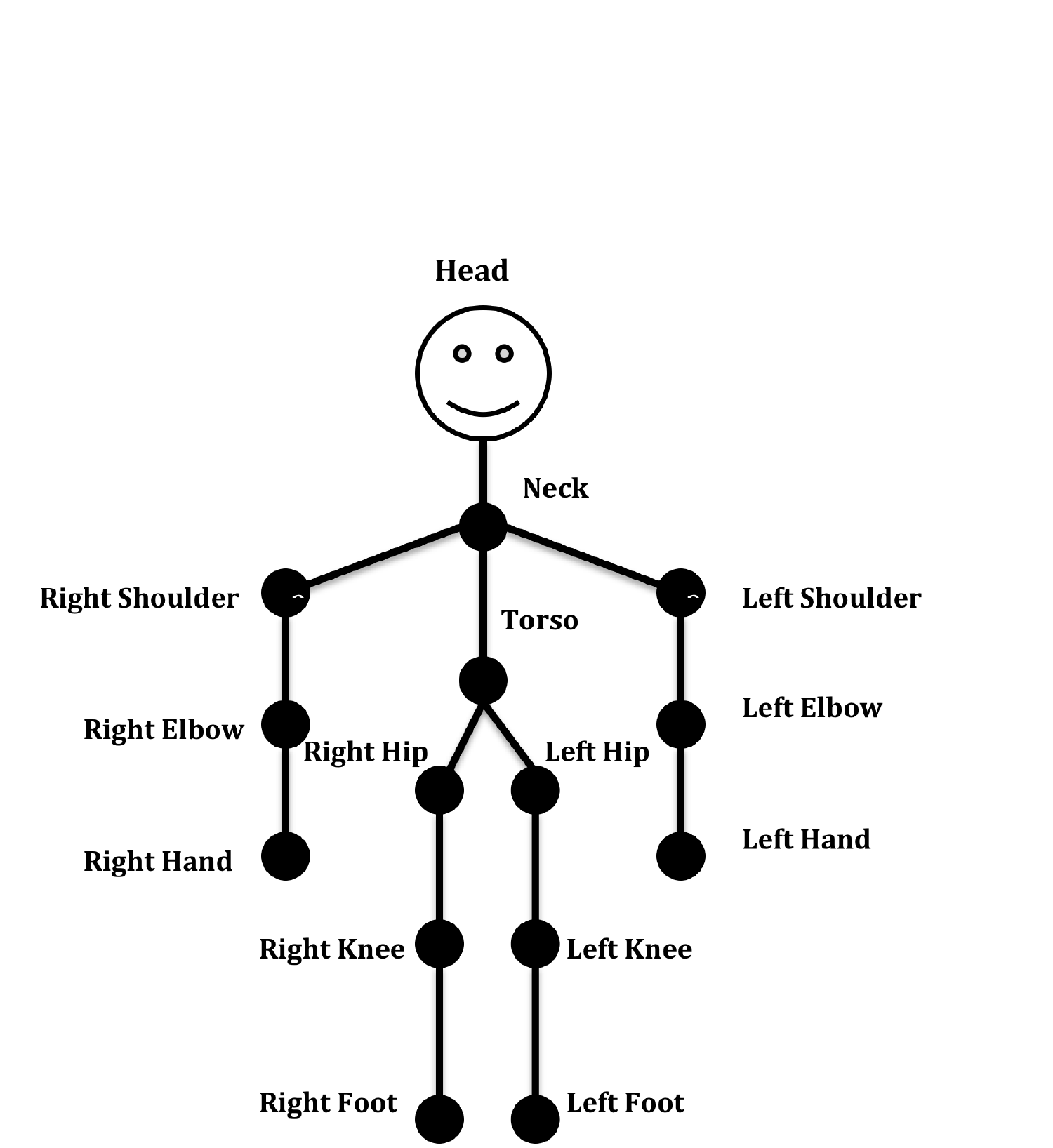}
  \caption{A representation of all connected joints which makes up one human posture.}
  \label{fig:skeleton_1}
 \end{figure}

Before entering the first-layer SOM, further preprocessing is applied to the input data. This is done because the distance and the capturing angle between the depth camera and the subjects performing the actions may vary. To overcome this in a robust way without having to rely only on the three neural network layers in the action recognition subsystem, the joint positions are re-scaled and transformed into an egocentric coordinate system. Thus, first the joint positions in each posture frame from the depth camera is re-scaled, i.e. made into a standard size.
Then the coordinates of the joint positions are transformed into a new and egocentric coordinate system located close to the torso joint of the skeleton, see Fig.~\ref{fig:skeleton_1}.

\begin{figure}[!h]
  \centering
   \includegraphics[width=1.10\textwidth]{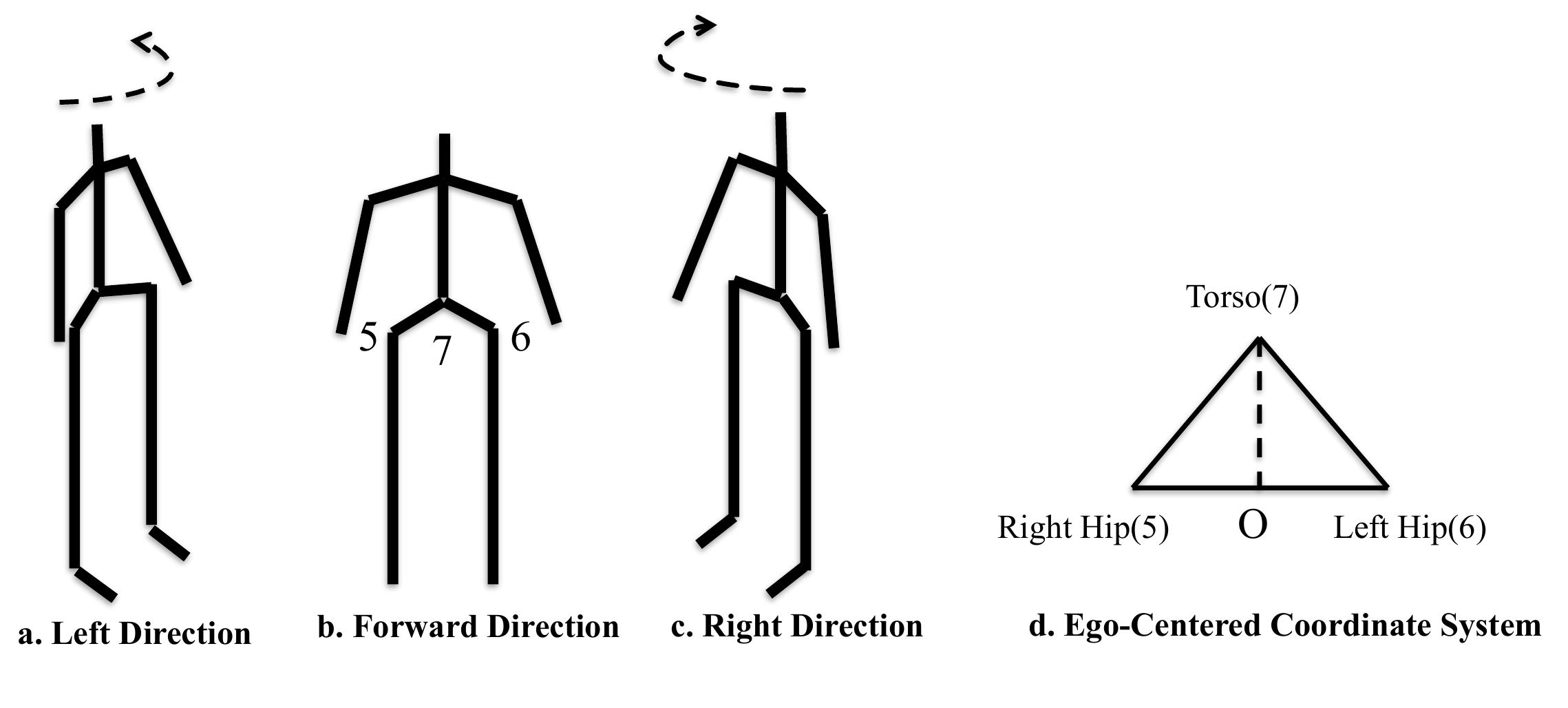}
  \caption{The actor with different orientations, turned to left(a), frontal direction(b), turned to right(c), and the joints used to calculate the egocentric coordinate system(d).}
  \label{fig:ego-cent}
 \end{figure}

To calculate the axes of the egocentric coordinate system, the joints 5 (Right Hip), 6 (Left Hip) and 7 (Torso) are used. As can be seen in Fig.~\ref{fig:ego-cent}, these joints constitute the vertices in a triangle and the projection $0$ of joint 7 on the side connecting joints 5 to 6 can be calculated. Then axes originating in the point $0$ along the line between the point $0$ and joint 7 and along the line between joints 5 and 6 can be selected together with an axis orthogonal to the triangle for the new coordinate system and a transformation matrix \citep{Craig} can be calculated, which allows all the joints to be expressed in egocentric coordinates.

\begin{figure}[!h]
  \centering
   \includegraphics[width=0.90\textwidth]{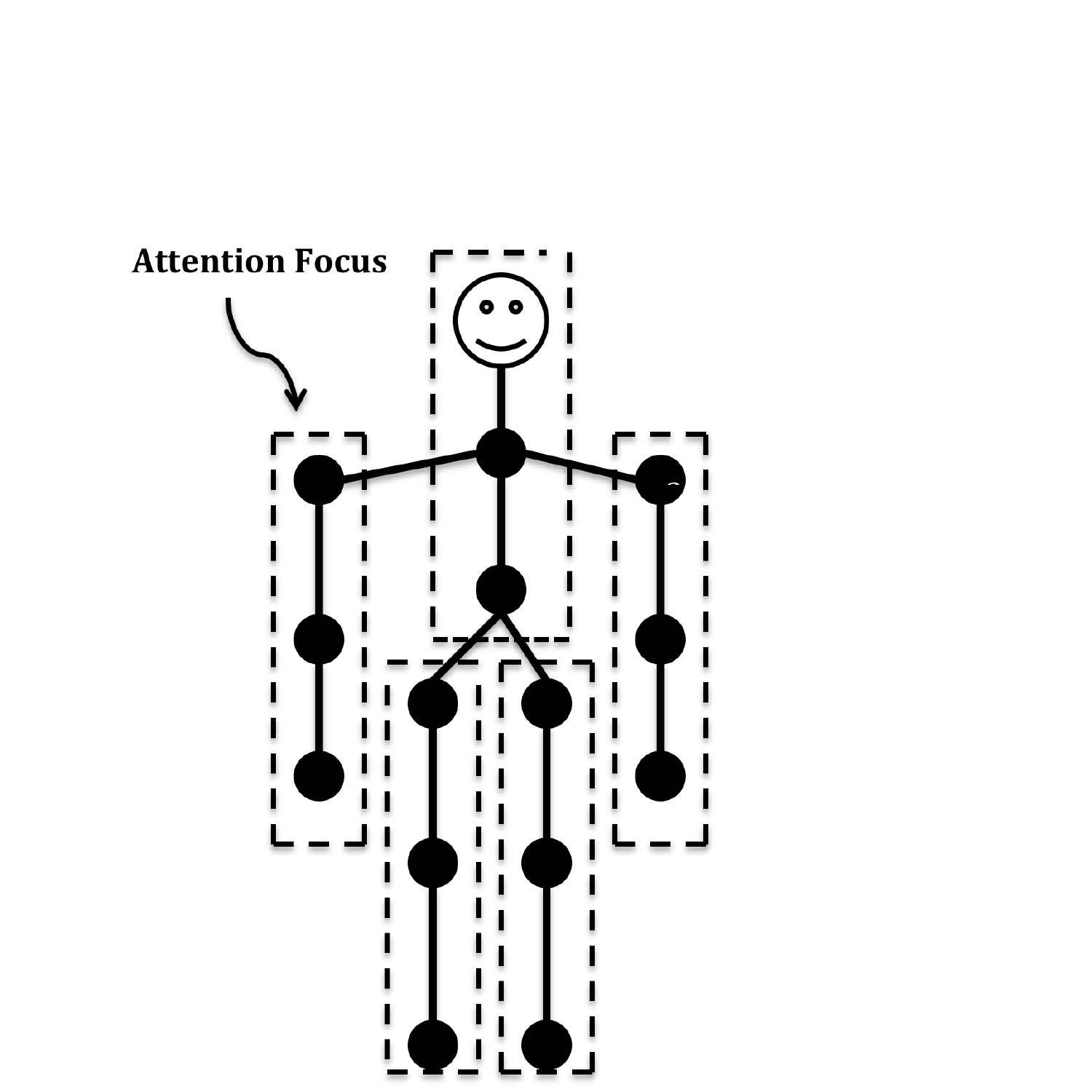}
  \caption{The division of the skeleton into five basic parts: Right Arm, Right Leg, Left Arm, Left Leg and Body.}
  \label{fig:Attention1}
 \end{figure}

Due to limitations in the visual field, time, and processing capacity, the entire input information cannot always be processed in real time \citep{Shariatpanahi}. By using attention mechanisms, performance can be improved, e.g. in a driving task \citep{Gharaee} or in the case of action perception in the current study. We applied an attentional mechanism to the part of the skeleton that exhibits the largest movements. In this way the influence of less relevant parts of the input data can be decreased whereas the influence of more relevant parts of the input data are increased. This is achieved by dividing the skeletons into five basic parts, Fig.~\ref{fig:Attention1}. The division is based on how actions are performed in a human body. The focus of attention is set to the moving part, which in the set of actions considered in this study is the right arm of the subjects. This is illustrated in Fig.~\ref{fig:Attention1}.\\
 
\noindent \textbf{First and Second Layer SOMs.} The first two layers of the action recognition subsystem consist of SOMs. The SOMs are trained using unsupervised learning to produce dimensionality reduced and discretized representations of their input spaces. These representations preserve the topology of their corresponding input spaces, which means that nearby parts of the network will respond to similar input patterns, reminiscent of the cortical maps found in mammalian brains. The SOMs will therefore generate a measure of similarity which is the founding property of a conceptual space \citep{Gardenfors4}. In other words, the map generated by a SOM can be seen as a conceptual space that is generated from the training data.

The topology-preserving property of SOMs is a consequence of the use of a neighbourhood function in the adaptation of the neuron responses, i.e. the adaptation strength is a decreasing function of the distance from the most activated neuron in the network. This also provides the SOM, and in the extension our action recognition system, with the ability to generalize learning to novel inputs, because similar inputs elicit similar activities in the SOM.

The SOM consists of an $I\times J$ grid of neurons with a fixed number of neurons and a fixed topology. Each neuron $n_{ij}$ is associated with a weight vector $w_{ij}\in{R}^n$ with the same dimensionality as the input vectors. All the elements of the weight vectors are initialized by real numbers randomly selected from a uniform distribution between 0 and 1.

At time $t$ each neuron $n_{ij}$ receives the input vector $x(t)\in{R}^n$.
The net input $s_{ij}(t)$ at time $t$ is calculated using the Euclidean metric:

\begin{equation}
s_{ij}(t)=||x(t) - w_{ij}(t)||
\end{equation}

The activity $y_{ij}(t)$ at time $t$ is calculated by using the exponential function:

\begin{equation}
y_{ij}(t)=e^{\frac{-s_{ij}(t)}{\sigma}}
\end{equation}

The parameter $\sigma$ is the exponential factor set to ${10^6}$  and $0 \leq {i} < I$, $0 \leq {j} < J$, ${i},{j}\in{N}$. The role of the exponential function is to normalize and increase the contrast between highly activated and less activated areas.

The neuron $c$ with the strongest activation is selected:

\begin{equation}
c=\mathrm {arg} \mathrm{ max}_{ij}y_{ij}(t)
\end{equation}

The weights $w_{ijk}$ are adapted by

\begin{equation}
w_{ijk}(t+1)=w_{ijk}(t)+\alpha(t)G_{ijc}(t)[x_k(t)-w_{ijk}(t)]
\end{equation}

The term $0 \leq \alpha(t) \leq 1$ is the adaptation strength, $\alpha(t) \rightarrow 0$ when $t \rightarrow \infty$. The neighbourhood function $G_{ijc}(t) = e^{-\frac{||r_c - r_{ij}||}{2\sigma^2(t)}}$ is a Gaussian function decreasing with time, and $r_c \in R^2$ and $r_{ij} \in R^2$ are location vectors of neurons $c$ and $n_{ij}$ respectively. \\

\begin{figure}[!t]
\centering
\includegraphics[width=3.50in]{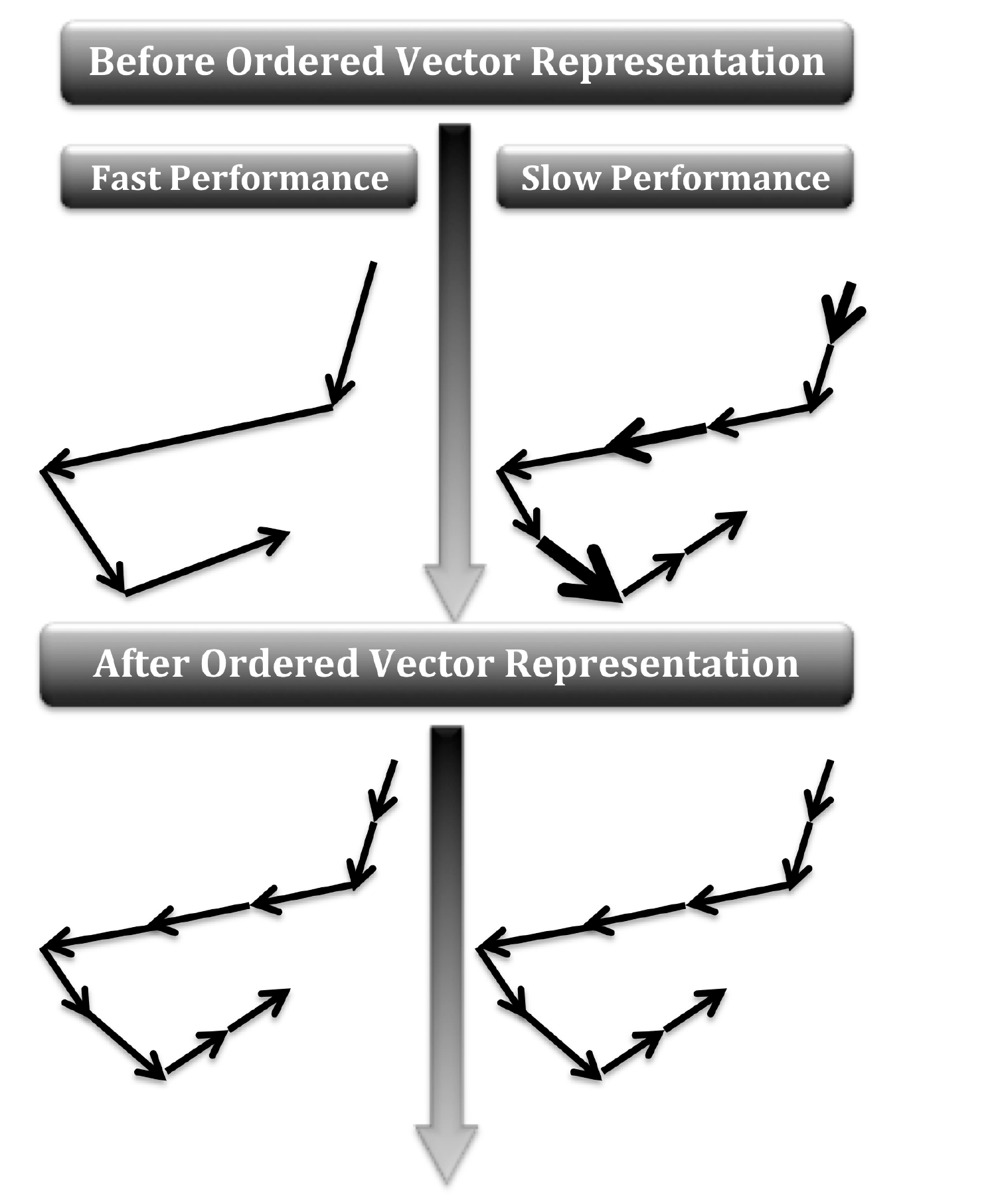}
\caption{Similar action sequences performed at different rates described as Fast Performance and Slow Performance will elicit similar activity traces in the first layer SOM. The Slow Performance represents a higher number of activated neurons in almost the same path. Moreover, there is repeated activation of the same neurons (shown by darker and thicker arrows) in the slow performance. After application of Ordered Vector Representation to both performances the same path with the same number of activated neurons is achieved.}
\label{supImp_nn}
\end{figure}

\noindent \textbf{Ordered Vector Representation.} The activity trace elicited in the first-layer SOM during the performance of an action is turned into a spatial representation and used as input to the second-layer SOM. This spatial representation is achieved by an ordered vector representation, thus keeping the order of the activity pattern elicited over time in the spatial representation while removing the temporal dependence. In this way time invariance is achieved. This is so because similar actions carried out at different speeds will be composed of similar sequences of significant postures or at least similar activity traces in the first layer SOM, although the number of input frames may vary in the sequences. Similar postures during an action will elicit similar activity in the first-layer SOM. Thus, similar actions performed at different rates will elicit similar activity traces in the first layer SOM, Fig.~\ref{supImp_nn}, i.e. they will start with a similar activity pattern that evolves and ends in a similar way during the action.

An activity trace elicited in the first layer SOM due to the performance of an action is made into an ordered vector representation  as follows: 

The length of the activity trace of an action $\Delta_{j}$ is calculated by 

\begin{equation}
  \Delta_{j} = \sum_{i=1}^{N-1} {{||P_{i+1}-P_{i}||}_2}
\end{equation}

The variable $N$ is the total number of centers of activity for action sequence $j$ and $P_{i}$ is the $i$th centre of activity in the same action sequence. 

Suitable lengths of segments to divide the activity trace for action sequence $j$ in the first-layer SOM are calculated by

\begin{equation}
  d_{j} = \Delta_{j}/N_{Max}
\end{equation}

The variable $N_{Max}$ is the longest path in the first-layer SOM elicited by the $M$ actions in the training data.

To achieve an ordered vector representation of the activity trace corresponding to an action, the activity trace is divided into $d_{j}$ segments, and the coordinates of the borders of these segments in the order they appear from the start to the end on the activity trace are composed into a vector. \\

\noindent \textbf{Supervised Layer.} The supervised layer constitutes the output layer of the action recognition subsystem. It receives the activity of the second-layer SOM as input and consists of an $I\times J$  grid with a fixed number of neurons and with a fixed topology. Each neuron $n_{ij}$ is associated with a weight vector  $w_{ij}\in{R}^n$. All the elements of the weight vector are initialized by real numbers randomly selected from a uniform distribution between 0 and 1.

At time \textit{t} each neuron $n_{ij}$ receives an input vector $x(t)\in{R}^n$.

The activity $y_{ij}(t)$ at time $t$ in the neuron $n_{ij}$ is calculated using the standard cosine metric:

\begin{equation}
y_{ij}(t)=\frac{x(t)\cdot w_{ij}(t)}{||x(t)||||w_{ij}(t)||}
\end{equation}

During the learning phase the weights $w_{ijl}$ are adapted by

\begin{equation}
w_{ijl}(t+1)=w_{ijl}(t)+\beta x_{l}(t)[y_{ij}(t) - d_{ij}(t)]
\end{equation}

The parameter  $\beta$ is the adaptation strength and $d_{ij}(t)$ is the desired activity for the neuron $n_{ij}$. The desired activity is the activity pattern in the output layer that corresponds to the unambiguous recognition of the ongoing action.

\subsection{Object Identification}
\label{sec:output}
The object identification subsystem simultaneously detects and tracks the present objects in an environment where several objects are present, and uses the obtained information together with the agent's joint positions to determine what particular object the action is applied to.\\
 
\noindent \textbf{Marker Tracker.} The present implementation bases the object tracking
 on markers attached to the objects. The marker tracker employed by our object identification subsystem is publicly available as a part of the distribution of the neural modelling framework Ikaros \citep{balkenius}. It provides the identities of the objects and their positions. The marker tracker could in principle be replaced by a more sophisticated object tracking mechanism that determines the identities of the objects by object recognition. The object involved in the action is determined by applying a proximity measure between the observed agent's hand and the present objects.\\
 
\noindent \textbf{Preprocessing.} In order to have both the joint positions of the agent and the positions of the objects expressed in the same coordinate system, the object's positions are also transformed into the same egocentric coordinate system, described above, as the agent.\\

\noindent \textbf{Proximity Measure.} Below, the method applied to determine which object the observed agent acts upon is described. In the present study, the actor performs five different actions: push the object; pull the object; put down the object; lift up the object; and point to the object. In all of these cases, the action involves one object. To make the task practical, especially because there are several ways of performing each of these actions, we limited the task to particular ways of performing the actions. Therefore, the action push occurs when the actor applies force to the object to move it horizontally away from herself, and pull is the same task but the force is applied in reverse direction to move the object horizontally closer to the subject. The actions put down and lift up are performed when the actor applies force to the object to move it vertically down or up. Finally, the action point is performed when the subject uses the index finger with a straight arm to show the object. 

We can conclude that when these actions are carried out, there is the same spatial trajectories for both the object and the agent's hand. Thus, by knowing the spatial trajectory of the hand we will also know the spatial trajectory of the object acted upon as well. Therefore, the information about the object's position while the actions are performed can be estimated from the position of the agent's hand. There is one exception to this, namely the action "point to the object". In that case the condition is a bit different. Although both the agent's hand and the object still have the same (stationary) motion trajectory, their relative locations in the environment can be different. This means that, unlike the other actions, while performing the action "point to the object", the object is not necessarily at the same position as the agent’s hand.. 

The task is to determine precisely which object among several present, the agent's action is directed towards. To this end, a proximity measure based on the Euclidean distance between the agent's hand and each of the present objects is calculated as follows:

\begin{equation}
  PM_{Object_{k}} = {||O_{pos}-S_{pos}||}_2 
\end{equation}
  
The parameter $PM_{Object_{k}}$ is the calculated proximity measure value for the object $k$. The object with the smallest proximity measure value is estimated to be the object the observed agent acts upon.

\section{Experiments}
\label{sec:exp}
\noindent We have evaluated our action recognition architecture by testing it online in real time. To this end, we recorded a set of human actions including objects performed by a human actor and used it to train the system. The actions used in this experiment are: 1. Push the Object, 2. Pull the Object, 3. Put the Object, 4. Lift the Object, 5. Point to the Object. The dataset used to train the system contains 12 samples of each of the 5 different action, i.e. in total 60 action samples. All action samples were performed by the same actor and three objects were present in the scene.

The action recognition subsystem, composed of a hierarchical neural network system was trained with randomly selected instances from the training set in two phases, the first to train the first-layer $30\times30$ neurons SOM, and the second to train the second layer $35\times35$ neurons SOM, and the supervised layer containing 5 neurons.

To test the action recognition architecture, a human performer carried out the same actions in real time in front of the system's Kinect sensor while the system was identifying the actions in online mode.   

\begin {table}[H]
\caption {The action recognition architecture perfectly learned the training samples for all of the actions.} \label{tab:title} 
\begin{tabular}{ |p{2.5cm}||p{2.8cm}|p{3cm}|p{2cm}| }
 \hline
 \multicolumn{4}{|c|}{\textbf{Training the System}}\\
 \hline
 Action   Names& Number of Train Samples &Number of Learned Samples& Training Results \\
 \hline
 Push Object   & 12    &12     & 100\% \\
 Pull Object     & 12    & 12    & 100\% \\
 Put Object     & 12   & 12     &100\%\\
 Lift Object     &12    & 12     &100\%\\
 Point to Object  &12    &12     &100\%\\
Total         &60    &60     &\textbf{100\%}\\
 \hline
\end{tabular}
\label{table1}
\end {table}

The performance of the action recognition architecture is shown in table.\ref{table1} and table.\ref{table2}. As can be seen in table.\ref{table1}, the system completely learned all training samples of all the actions. In the online test experiment, table.\ref{table2}, carried out in real time, the action recognition architecture performed quite excellent. As can be seen, the recognition rates of all the actions are quite high. However, the action recognition architecture performed slightly worse when it comes to the action Point to the Object. The reason for this could be that the nature of this action is more complicated than the others. In fact pointing is composed of basically one gesture which means that it is a repetition of similar postures during the whole action, in which the arm is located in a fixed position and has the same gesture. This increases the probability of overlapping with other actions in the first-layer SOM representation in the action recognition subsystem, and the influence of noise.

\begin {table}[H]
\caption {The performance of the online action recognition architecture when tested in real time with actions involving objects.} \label{tab:title} 
\begin{tabular}{ |p{2cm}||p{2cm}|p{2cm}|p{2cm}|p{2cm}|  }
 \hline
 \multicolumn{5}{|c|}{\textbf{Generalization Test Performance}}\\
 \hline
 Action     Names& Number of Test Samples &Number of Recognized Samples&Action Recognition Results &Object Detection Results\\
 \hline
 Push Object   & 9    &9     & 100\% &100\%\\
 Pull Object     & 9    & 9    & 100\% &100\%\\
 Put Object     & 9   & 8     & 88.9\% &100\%\\
 Lift Object     &9    & 8    & 88.9\% &100\%\\
 Point Object  &9    & 7    & 77.8\% &100\%\\
Total         &45  &41  & \textbf{91.1\%} &\textbf{100\%}\\
 \hline
\end{tabular}
\label{table2}
\end {table}

\begin{figure}[H]
\centering
\includegraphics[width=5.00in]{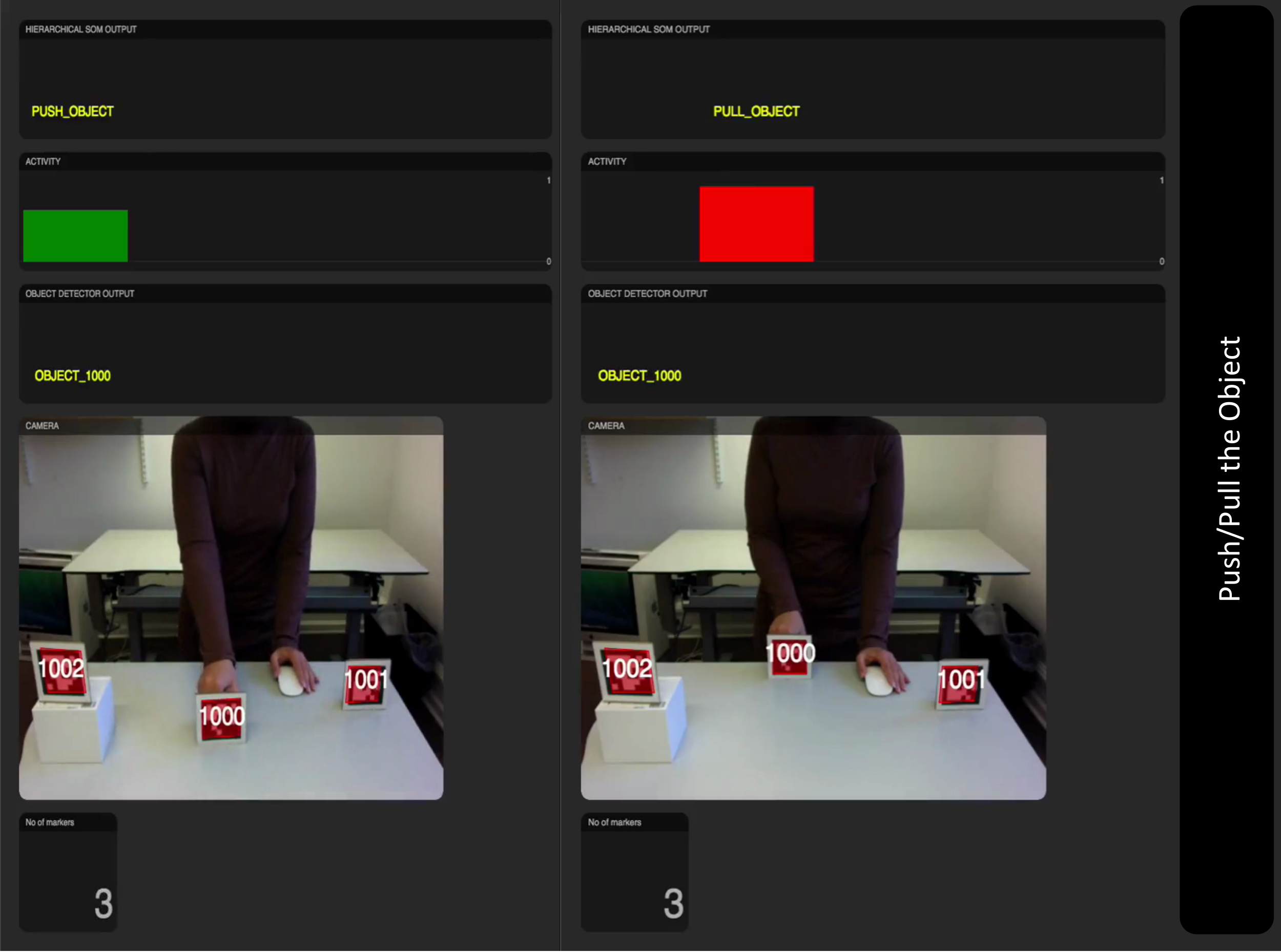}
\caption{The performance of the online action recognition architecture when tested in real time with actions involving objects such as Push and Pull the Object. The first row of the figure shows the system's recognition results as the name of the recognised action and the second row represents the activities of the corresponding neurons in the third layer neural network. It is shown that the neuron corresponding to the performed action has the largest activation value. The third row of the figure shows the object detection result by the index of the object. The number of all existing objects  detected by the system is also shown in the left bottom of the figure.}
\label{Demo1}
\end{figure}

\begin{figure}[H]
\centering
\includegraphics[width=5.00in]{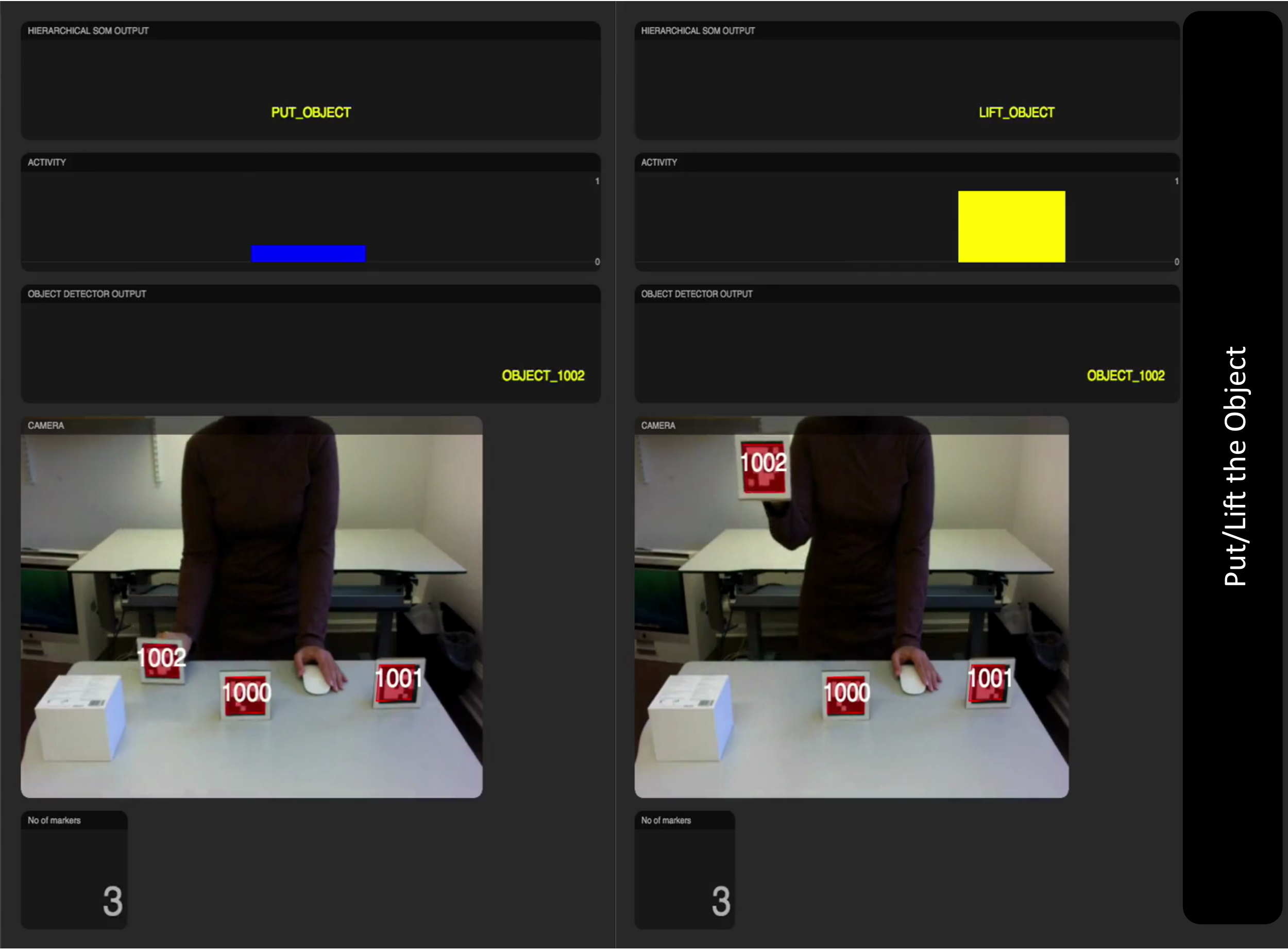}
\caption{The performance of the online action recognition architecture when tested in real time with actions involving objects such as Put and Lift the Object. The first row of the figure shows the system's recognition results as the name of the recognised action and the second row represents the activities of the corresponding neurons in the third layer neural network. It is shown that the neuron corresponding to the performed action has the largest activation value. The third row of the figure shows the object detection result by the index of the object. The number of all existing objects  detected by the system is also shown in the left bottom of the figure.}
\label{Demo2}
\end{figure}

\begin{figure}[H]
\centering
\includegraphics[width=5.00in]{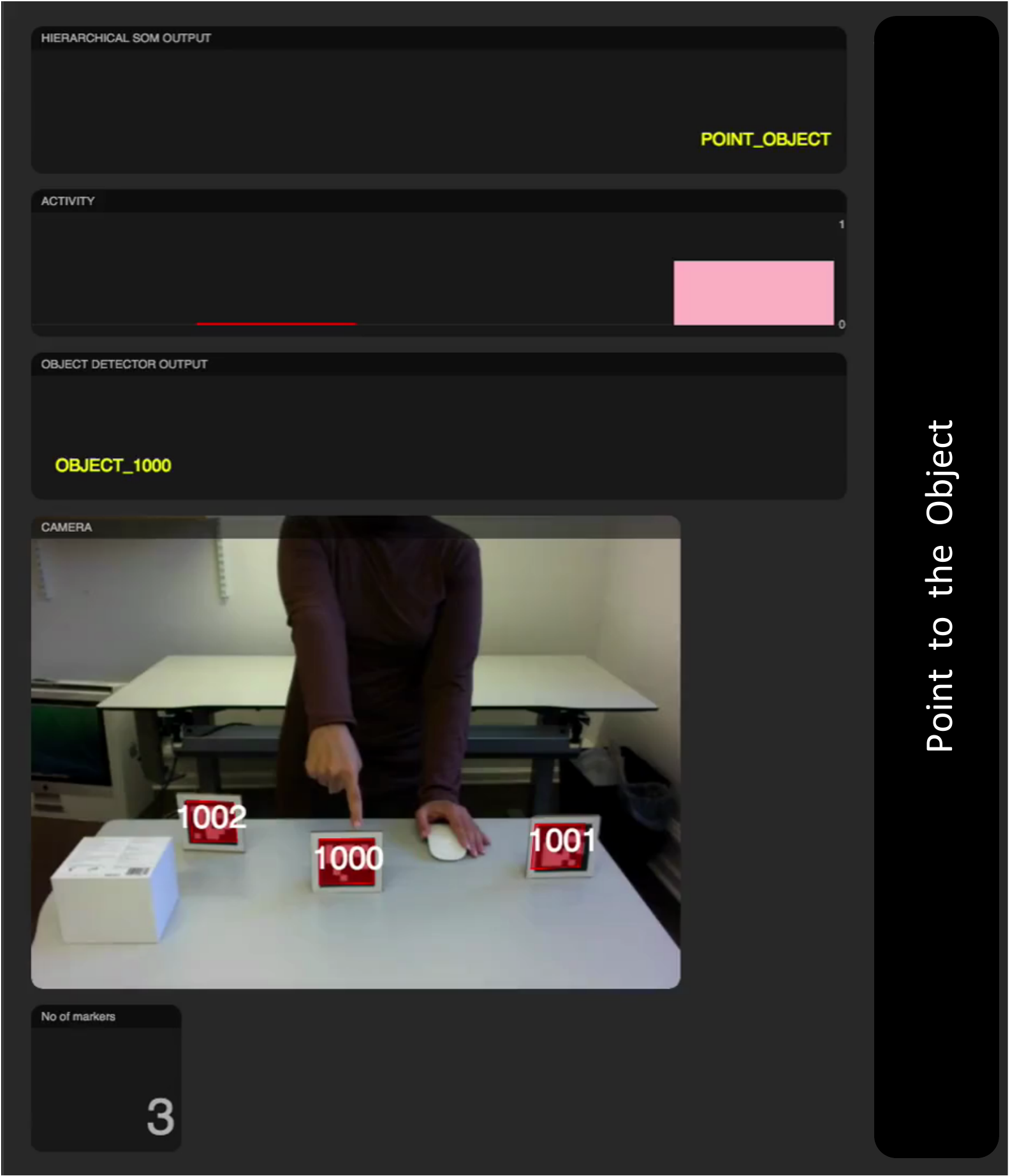}
\caption{The performance of the online action recognition architecture when tested in real time with the action Point to the Object. The first row of the figure shows the system's recognition results as the name of the recognised action and the second row represents the activities of the corresponding neurons in the third layer neural network. It is shown that the neuron corresponding to the action Point to the Object has the largest activation value. The third row of the figure shows the object detection result by the index of the object.The number of all existing objects  detected by the system is also shown in the left bottom of the figure.}
\label{Demo3}
\end{figure}

As mentioned earlier we implemented the online version of our architecture in the modeling framework Ikaros \citep{balkenius}. The experiments are carried out with this implementation and the results were filmed and a demo movie was created, which is available through the webpage \cite{Johnsson3}. In Fig.~\ref{Demo1}, Fig.~\ref{Demo2} and Fig.~\ref{Demo3} snapshots from the movie demonstrating online action recognition in real time are shown for the actions used in our experiments.

\section{Discussion}

\noindent We have presented a system for online action recognition in real time that merges the information analyses of two subsystems running in parallel. The first subsystem receives input data from a depth camera and recognizes what action is performed by using a hierarchical SOM system that analyses the spatial trajectories of the agent's movements. The SOMs in the first and second layers enable compressed representations of postures and agent movements, thus reducing the high dimensionality of the input data. The second subsystem detects objects and determines which object the agent acts upon by applying a proximity measure.

\noindent The first-layer SOM of the first subsystem develops an ordered representation of postures, and during an action, an activity trajectory characteristic to the particular ongoing action is elicited in this SOM. The activity trajectory elicited in the first-layer SOM during an action is arranged into an ordered vector representation before it is received by the second-layer SOM. This ordered vector representation provides a way of handling a varying number of activations in the first-layer SOM that result when an action is carried out at different rates. This means that the system achieves time invariance to actions.

Earlier versions of our architecture have been tested with data sets from the MSRAction3D \citep{MSR} in offline experiments of several studies. In one study \citep{Gharaee3}, we used a subset of 276 action sequences to determine how well these actions could be classified on the basis of sequences of joint positions. We obtained 83\% correct classifications with the test set containing samples not used for training. In another study using these data, our architecture classified the actions not only on the basis of sequences of joints positions (postures), but also on their first and second order dynamics representing joints velocity and acceleration.This resulted in an improvement in the performance of action classification \citep{Gharaee4}. With variations, the architecture presented in this study has also been tested in the experiments on a completely different data set of 2D movies as input in \cite{buonamente5}, in offline experiments. 

An online version of our action recognition architecture was tested and presented in \cite{Gharaee2}. In those experiments, the action recognition system was successfully tested with both actions not involving objects as well actions that do involve objects. However, the system did not contain the second information processing stream for identifying objects  so it couldn't determine whether there actually was an object involved in the action, or if it was just a pantomime. Neither could it determine which object the agent acted upon, if present, in the case of actions involving objects. In the experiments with our former version of the online action recognition system only the actor's spatial trajectory was used for the recognition.

\cite{Ellis} and \cite{Vieira} also present results for online action recognition experiments. In \cite{Vieira}, a subset of MSR data manner actions is used, and in \cite{Ellis} the proposed method is tested online by using a new set of manner actions. In both cases, only actions that involve body movements without objects are considered. This limitation has been addressed in the present study since we used actions that involve objects. Thus when an action is performed the resulting state shows the changes in the world, for example, the movement of a cup. The information of the resulting state is also extracted in order to identify the target object among several objects present in the environment. 

The performance results of our architecture for recognition of actions involving objects is quite good, and almost in all cases of our experiments the system is capable of successfully recognizing the performed action, as well as detecting which objects the actor acts upon. 

We believe that our work contributes to the challenge of biologically inspired cognitive architectures \cite{Samsonovich}. The system we have implemented borrows several ideas from the studies of biological motion within psychology. The system relates several of the desiderata that  \cite{Samsonovich} lists, in particular, accepting the system as a partner, attention and intentionality, human compatibility, and sense making. The typical output of the action recognition system is a verb plus some information about an object acted on (for example change of position). This output can be fed into a human-robot communication system of the type implemented in \cite{Mealier} and it can become an important component in generating the sentences that are expressed by the robot, for example when describing an event.

In future work, we plan to extend our experimentation to a larger group of actions with different numbers of agents and objects. One of our motivations for a SOM based approach for action recognition is to apply a method for segmenting each sequence of an action so that the system will have no more need to receive a signal of the start/end of the actions. To this end, we will rely on the key postures of each sequence of the actions which are extracted in the first SOM of our hierarchical architecture. Another motivation is to make it possible to internally simulate the likely continuation of partly seen actions. This can be done by employing Associative Self-Organizing Maps (A-SOMs) \citep{johnsson1}, and have been investigated by using 2D movies as input in studies presented in \cite{buonamente4}.

\section*{References}

\bibliography{mybibfile}

\end{document}